\documentclass[letterpaper, 10 pt, journal, twoside]{IEEEtran}

\IEEEoverridecommandlockouts    

\usepackage{graphics} 
\usepackage{epsfig} 
\usepackage{mathptmx} 
\usepackage{times} 
\usepackage{amsmath} 
\usepackage{amssymb} 
\usepackage{nidanfloat}
\usepackage{multirow}
\usepackage{float}
\usepackage{array}
\usepackage{color}
\usepackage[bookmarks=false]{hyperref}
\usepackage{threeparttable}
\usepackage{etoolbox}
\usepackage{ifpdf}
\usepackage[utf8]{inputenc}
\makeatletter
\let\NAT@parse\undefined
\makeatother

\makeatletter
\newcommand*{\inlineequation}[2][]{%
  \begingroup
    \refstepcounter{equation}%
    \ifx\\#1\\%
    \else
      \label{#1}%
    \fi
    \relpenalty=10000 %
    \binoppenalty=10000 %
    \ensuremath{%
      #2%
    }%
    ~\@eqnnum
  \endgroup
}
\makeatother

\usepackage{hyperref}  

\newcommand{\bhline}[1]{\noalign{\hrule height #1}}
\def\figref#1{{Fig.~}\ref{#1}}
\def\tabref#1{{Tab.~}\ref{#1}}

\newcommand{\addref}{\color[rgb]{0., 0., 0.}}
\newcommand{\add}[1]{\textcolor[rgb]{0., 0., 0.0}{#1}}

\newcommand{\unnecessary}[1]{}

\let\mybibitem\bibitem
\renewcommand{\bibitem}[1]{
  \ifstrequal{#1}{sturmBenchmarkEvaluationRGBD2012a}
    {\addref \mybibitem{#1}}
    {\ifstrequal{#1}{delmericoAreWeReady2019}
        {\addref \mybibitem{#1}}
        {\ifstrequal{#1}{antoniniBlackbirdUAVDataset2020a}
            {\addref \mybibitem{#1}}
            {\ifstrequal{#1}{qinGeneralOptimizationbasedFramework2019}
                {\addref \mybibitem{#1}}
                {\ifstrequal{#1}{camposORBSLAM3AccurateOpenSource2020}
                    {\addref \mybibitem{#1}}
                    {\ifstrequal{#1}{bescosEmptyCitiesDynamicObjectInvariant2020}
                        {\addref \mybibitem{#1}}
                        {\ifstrequal{#1}{bescosDynaSLAMIITightlyCoupled2020}
                            {\addref \mybibitem{#1}}
                            {\ifstrequal{#1}{qiuTracking3DMotion2019}
                                {\addref \mybibitem{#1}}
                                {\ifstrequal{#1}{sabzevariMultibodyMotionEstimation2016}
                                    {\addref \mybibitem{#1}}
                                    {\ifstrequal{#1}{juddMultimotionVisualOdometry2018}
                                        {\addref \mybibitem{#1}}
                                        {\ifstrequal{#1}{bescosDynaSLAMIITightlyCoupled2020}
                                            {\addref \mybibitem{#1}}
                                            {\color{black}\mybibitem{#1}}
                                        }
                                    }
                                }
                            }
                        }
                    }
                }
            }
        }
    }
}

\title{\LARGE \bf
VIODE: A Simulated Dataset to Address the Challenges of Visual-Inertial Odometry in Dynamic Environments
}

\author{Koji Minoda$^{1}$, Fabian Schilling$^{2}$, Valentin Wüest$^{2}$, Dario Floreano$^{2}$, and Takehisa Yairi$^{1}$
\thanks{Manuscript received: October, 15, 2020; Revised January, 10, 2021; Accepted January, 31, 2021.}
\thanks{This paper was recommended for publication by Editor Javier Civera upon evaluation of the Associate Editor and Reviewers' comments.}
\thanks{$^{1}$Koji Minoda and Takehisa Yairi are with the Artificial Intelligence Laboratory, Department of Aeronautics and Astronautics, University of Tokyo, Tokyo, Japan.
{\tt\footnotesize koji.m.minoda@gmail.com}}%
\thanks{$^{2}$Fabian Schilling, Valentin Wüest, and Dario Floreano are with the Laboratory of Intelligent Systems, École Polytechnique Fédérale de Lausanne, Lausanne, Switzerland.}%
\thanks{Digital Object Identifier (DOI): see top of this page.}
}

\begin{document}
\maketitle

\begin{abstract}
Dynamic environments such as urban areas are still challenging for popular visual-inertial odometry (VIO) algorithms.
Existing datasets typically fail to capture the dynamic nature of these environments, therefore making it difficult to quantitatively evaluate the robustness of existing VIO methods.
To address this issue, we propose three contributions: firstly, we provide the VIODE benchmark, a novel dataset recorded from a simulated UAV that navigates in challenging dynamic environments.
The unique feature of the VIODE dataset is the systematic introduction of moving objects into the scenes.
It includes three environments, each of which is available in four dynamic levels that progressively add moving objects. 
The dataset contains synchronized stereo images and IMU data, as well as ground-truth trajectories and instance segmentation masks.
Secondly, we compare state-of-the-art VIO algorithms on the VIODE dataset and show that they display substantial performance degradation in highly dynamic scenes.
Thirdly, we propose a simple extension for visual localization algorithms that relies on semantic information.
Our results show that scene semantics are an effective way to mitigate the adverse effects of dynamic objects on VIO algorithms.
Finally, we make the VIODE dataset publicly available at \url{https://github.com/kminoda/VIODE}.

\begin{IEEEkeywords}
Data Sets for SLAM, Visual-Inertial SLAM, Aerial Systems: Perception and Autonomy
\end{IEEEkeywords}
\end{abstract}

\section{Introduction}

\IEEEPARstart{V}{ision-based} localization in dynamic environments is an important and challenging task in robot navigation.
Camera images can provide a rich source of information to estimate the pose of a robot, especially in environments such as indoor and urban areas where GNSS information may be unreliable or entirely unavailable. 
In such environments, however, dynamic objects such as humans, vehicles, or other mobile robots often coexist in the same workspace.
These dynamic objects can be detrimental to vision-based algorithms since they commonly use the assumption of a static world in which they self-localize.
Breaking the static-world assumption may
result in large estimation errors since the algorithms cannot distinguish between dynamic objects and static ones.

One possible way to remove these errors is to introduce additional proprioceptive information, such as data from inertial measurement units (IMUs), into the estimation pipeline.
Unlike a camera that measures the environment, proprioceptive sensors measure the internal state of the robot, which is usually not affected by the environment. 
Thus, visual-inertial odometry (VIO) has the potential to perform more robustly compared to purely visual odometry (VO).
However, as we show in this study, adding inertial information does not necessarily guarantee robustness, and further information is required to obtain a reliable state estimation algorithm.

Despite the importance of visual-inertial fusion for robust localization in dynamic scenarios, there are no publicly available datasets which could be used as a benchmark of VIO algorithms in dynamic environments.
While a considerable amount of VIO datasets are proposed, most of them contain only a few dynamic objects.
While there are VIO algorithms that aim to perform robustly in dynamic environments~\cite{muVisualNavigationFeatures2020, baiPerceptionaidedVisualInertialIntegrated2020}, it is difficult to quantitatively compare them due to the lack of a specific benchmark.

\begin{figure}[t]
    \centering
    \includegraphics[width=\columnwidth]{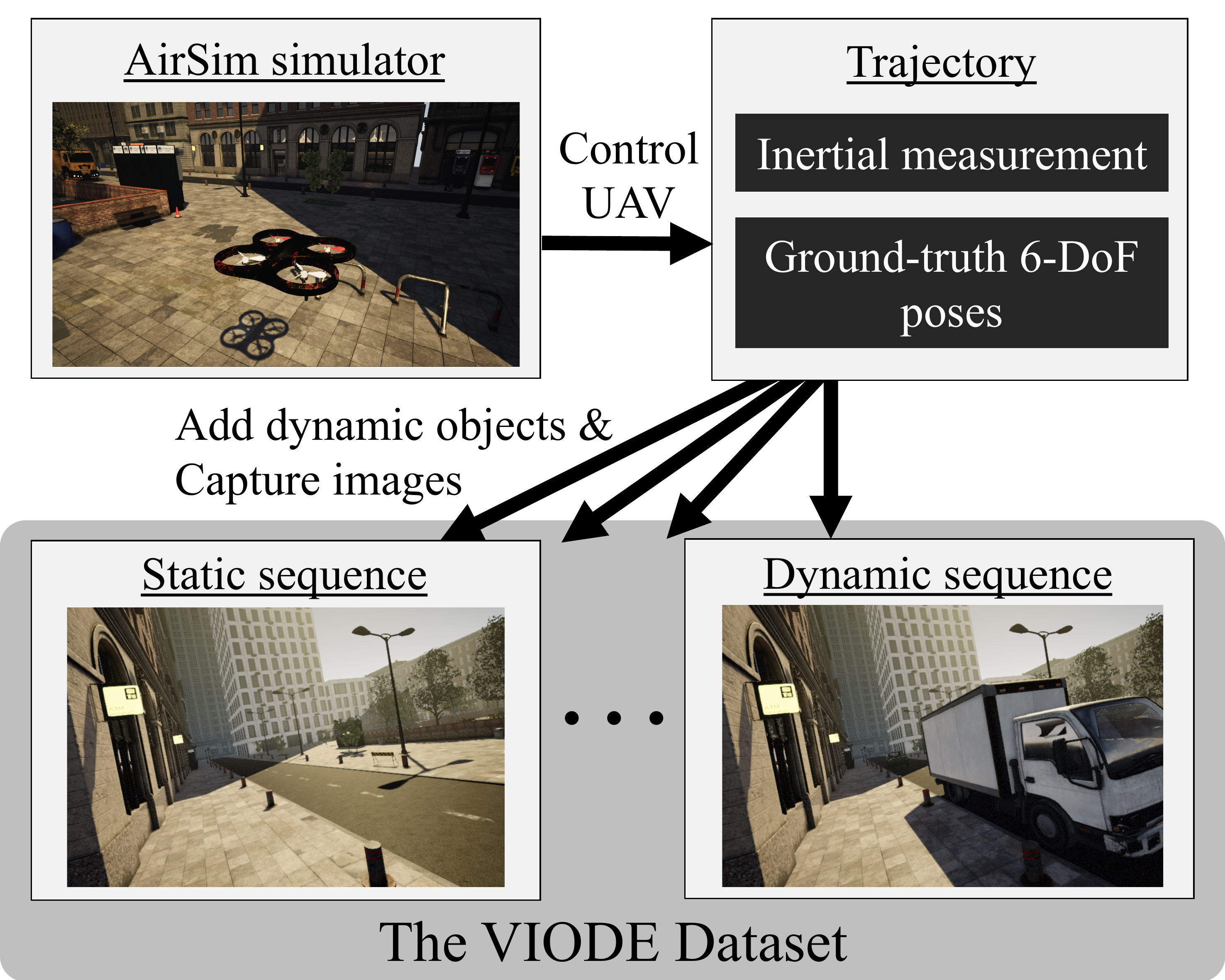}
    \caption{\footnotesize The VIODE dataset is created using a photorealistic UAV simulator to which we systematically add moving objects while keeping trajectory, lighting conditions, and static objects the same -- a setup which would be prohibitively difficult to achieve with real-world data. Each environment in the dataset contains four sequences with the same trajectory and IMU data, but with increasing levels of dynamic objects.}
    \label{fig:catchy}
\end{figure}

This paper presents a novel dataset called VIODE (VIO dataset in Dynamic Environments) -- a benchmark for assessing the performance of VO/VIO algorithms in dynamic scenes.
The environments are simulated using AirSim~\cite{shahAirSimHighFidelityVisual2017}, which is a photorealistic simulator geared towards the development of perception and control algorithms.
The unique advantage of VIODE over existing datasets lies in the systematic introduction of dynamic objects at increasing numbers and in different environments (see \figref{fig:catchy}).
In each environment, we use the same trajectory of the aerial vehicle to create data sequences with an increasing number of dynamic objects.
Therefore, VIODE users can isolate the effect of the dynamic level of the scene on the robustness of vision-based localization algorithms.
Using VIODE, we show that the performance of state-of-the-art VIO algorithms degrades as the scene gets more dynamic. 

Our dataset further contains ground-truth instance segmentation masks, since we believe that semantic information will be useful for researchers and engineers to develop robust VIO algorithms in dynamic scenes.
In order to assess the effects of semantic information in algorithms operating on the VIODE dataset, we develop a method incorporating semantic segmentation into VINS-Mono~\cite{qinVINSMonoRobustVersatile2018}.
We refer to this method as VINS-Mask in our research.
VINS-Mask masks out the objects which are assumed to be dynamic.
Using our dataset, we demonstrate that VINS-Mask outperforms the existing state-of-the-art algorithms.
Though VINS-Mask uses ground-truth segmentation labels provided by AirSim, these results indicate the potential of the utilization of scene semantics in dynamic environments.

In summary, the main contributions of our work are:
\begin{itemize}
    \item We propose a novel simulated visual-inertial dataset (VIODE) where we systematically add dynamic objects to allow benchmarking of the robustness of VO and VIO algorithms in dynamic scenes.
    \item Using VIODE, we experimentally show performance degradation of state-of-the-art VIO algorithms caused by dynamic objects.
    \item We show that incorporation of semantic information, implemented as a state-of-the-art VIO algorithm with semantic segmentation, can alleviate the errors caused by dynamic objects.
\end{itemize}
\section{Related Work}
\begin{table*}[htbp]
\vspace{0.5cm}
    \renewcommand\arraystretch{1.2}
    \centering
    \caption{\textsc{comparison of existing datasets and ours}}
\begin{threeparttable}[h]
\begin{tabular}{l>{\raggedright}p{0.03\linewidth}>{\raggedright\arraybackslash}p{0.08\linewidth}>{\raggedright\arraybackslash}p{0.15\linewidth}>{\raggedright\arraybackslash}p{0.2\linewidth}>{\raggedright\arraybackslash}p{0.09\linewidth}>{\raggedright\arraybackslash}p{0.1\linewidth}}
\bhline{0.8pt}
Dataset name & Year & Carrier & Sensor setup & Ground-truth trajectory & Environment & Dynamic level\\ \hline

KITTI \cite{geigerVisionMeetsRobotics2013} & 2012 & Car & Stereo, IMU, LiDAR & Fused IMU, GNSS & Outdoors & Low\\ 

Malagan Urban \cite{blanco-claracoMalagaUrbanDataset2014} & 2014 & Car & Stereo, IMU & GNSS & Outdoors & Mid \\ 

UMich NCLT \cite{carlevaris-biancoUniversityMichiganNorth2016} & 2016 & Segway & Omni, IMU, LiDAR & Fused GNSS, IMU, LiDAR & In-/outdoors & Low \\ 

EuRoC MAV \cite{burriEuRoCMicroAerial2016} & 2016 & UAV & Stereo, IMU & MoCap & Indoors & Low \\ 

Zurich Urban \cite{majdikZurichUrbanMicro2017a}& 2017 & UAV & Monocular, IMU & Fused camera, Google Street View & Outdoors & Mid \\ 

PennCOSYVIO \cite{pfrommerPennCOSYVIOChallengingVisual2017} & 2017 & Handheld & Stereo, IMU & AprilTag markers & In-/outdoors & Low \\ 

TUM VI \cite{schubertTUMVIBenchmark2018} & 2018 & UAV & Stereo, IMU & MoCap (partial) & in-/outdoors & Low \\ 

ADVIO \cite{cortesADVIOAuthenticDataset2018}& 2018 & Handheld & Monocular, ToF, IMU & IMU with manual pose fix & In-/outdoors & High \\ 

Urban@CRAS \cite{gasparUrbanCRASDataset2018b} & 2018 & Car & Stereo, IMU, LiDAR & GNSS (partial) & Outdoors & Mid \\  

Oxford Multimotion \cite{juddOxfordMultimotionDataset2019} & 2019 & Handheld & Stereo, RGBD, IMU & MoCap & Indoors & Mid \\

KAIST Urban \cite{jeongComplexUrbanDataset2019}& 2019 & Car & Stereo, IMU & Fused GNSS, Fiber optic gyro, encoder, LiDAR & Outdoors & Mid  \\ 

OIVIO \cite{kasperBenchmarkVisualInertialOdometry2019} & 2019 & Handheld & Stereo, IMU & MoCap (partial), ORB-SLAM2 (partial)   & In-/outdoors & Low \\ 

\add{UZH-FPV Drone Racing \cite{delmericoAreWeReady2019}} & 2019 & UAV & Stereo, IMU, event & Laser tracker & In-/outdoors & Low\\ 

UMA-VI \cite{zuniga-noelUMAVIDatasetVisual2020} & 2020 & Handheld & Stereo, IMU & Camera (partial) & In-/outdoors & Low \\ 

\add{Blackbird UAV \cite{antoniniBlackbirdUAVDataset2020a}} & 2020 & UAV & Stereo\tnote{\add{*}}, IMU, depth\tnote{\add{*}}, segmentation\tnote{\add{*}} & MoCap & In-/outdoors & Low \\

\textbf{VIODE (ours)}  & 2021 & UAV (simulation) & Stereo\tnote{\add{*}}, IMU\tnote{\add{*}}, segmentation\tnote{\add{*}} & Simulation & In-/outdoors & High\\ \bhline{0.8pt}

\end{tabular}
\begin{tablenotes}
\item[\add{*}] \add{Simulated sensor}
\end{tablenotes}
\end{threeparttable}
\label{tab:datasets_comparison}
\end{table*}

In this section, we briefly outline related work in three areas. 
Firstly, we summarize existing VIO datasets and their shortcomings.
Secondly, we discuss the state-of-the-art of VIO algorithms.
Finally, we highlight methods for robust vision-based localization in dynamic environments.

\subsection{VIO datasets}
Existing VIO datasets are diverse, as they span various carriers, environments, and sensor setups. However, they are not sufficient to assess the robustness of VIO in dynamic environments on 6-DoF trajectories.

Most of the datasets that are recorded on UAV or handheld rigs contain only few moving objects.
For instance, widely used VIO benchmarks such as the EuRoC MAV dataset~\cite{burriEuRoCMicroAerial2016} only contain a small number of people moving in indoor scenes and thus the scenes are mostly static.
Thus they are not suitable for benchmarks for systematically studying performance in environments with moving objects.

Outdoor datasets tend to contain more dynamic objects.
The TUM VI dataset~\cite{schubertTUMVIBenchmark2018} and the Zurich Urban dataset~\cite{majdikZurichUrbanMicro2017a} contain outdoor sequences where sporadically moving vehicles and people appear.
KITTI~\cite{geigerVisionMeetsRobotics2013}, Urban@CRAS~\cite{gasparUrbanCRASDataset2018b}, and KAIST Urban~\cite{jeongComplexUrbanDataset2019} are recorded from driving cars in urban areas. 
These datasets, especially the latter two, contain several moving vehicles.
However, the field of view in these datasets is still mainly filled with static objects and are thus not challenging enough to benchmark robustness in dynamic environments.
The Oxford Multimotion Dataset~\cite{juddOxfordMultimotionDataset2019} is an indoor dataset which contains dynamic objects. 
Their dataset aims at providing a benchmark for motion estimation of moving objects as well as vehicle self-localization.
However, the scenes do not contain environments typically encountered by robots such as UAVs.
Furthermore, these datasets do not cover the most challenging dynamic scenes which can occur in real-world applications of vision-based localization.

To the best of our knowledge, the ADVIO dataset~\cite{cortesADVIOAuthenticDataset2018} contains the most challenging scenes in terms of dynamic level among the existing VIO datasets.
However, their work does not contain quantitative information on the dynamic level.
Moreover, unlike our VIODE dataset, the effect of dynamic objects cannot be isolated on the robustness of VIO.

In conclusion, the existing VIO datasets are not suitable to systematically benchmark the robustness of VIO methods in the presence of dynamic objects. 
The VIODE dataset seeks to enhance the development of localization in dynamic scenes by providing a challenging, quantitative, and high-resolution benchmark.
We achieve this goal with a systematic configuration of dynamic objects and an evaluation of dynamic levels in each scene.
\tabref{tab:datasets_comparison} provides a comparison of existing datasets and the VIODE dataset.

\subsection{VIO algorithms}
Here, we provide an overview of existing VIO algorithms, including VINS-Mono~\cite{qinVINSMonoRobustVersatile2018} and ROVIO~\cite{bloeschRobustVisualInertial2015} which we benchmark on the VIODE dataset.

Common VIO algorithms can be classified as either filter-based or optimization-based.
MSCKF~\cite{mourikisMultiStateConstraintKalman2007} and ROVIO~\cite{bloeschRobustVisualInertial2015} are both filter-based algorithms that fuse measurements from cameras and IMUs using an extended Kalman filter.
On the contrary, optimization-based approaches utilize energy-function representations for estimating 6-DoF poses.
OKVIS~\cite{leuteneggerKeyframebasedVisualInertial2015} utilizes non-linear optimization and corner detector.
VINS-Mono~\cite{qinVINSMonoRobustVersatile2018} 
\add{and VINS-Fusion~\cite{qinGeneralOptimizationbasedFramework2019} are} optimization-based algorithms that contain a loop detection and closure module and, as a result, achieve state-of-the-art accuracy and robustness.
\unnecessary{
All of these algorithms provide open-source implementations of their algorithms as Robot Operating System (ROS) compatible packages.
}
\add{
Furthermore, the recently proposed ORB-SLAM3~\cite{camposORBSLAM3AccurateOpenSource2020} is an optimization-based visual-inertial SLAM algorithm which achieves high performance and versatility.
}
In \cite{delmericoBenchmarkComparisonMonocular2018} the authors provide a comprehensive comparison among the existing open-sourced monocular VIO algorithms.
Their survey highlights that VINS-Mono and ROVIO are the two best-performing methods among the existing monocular VIO algorithms in terms of accuracy, robustness, and computational efficiency.
Thus, in this study, we compare these two algorithms on the VIODE dataset.

\subsection{Vision-based localization in dynamic environments}
In the context of vision-based localization algorithms such as visual odometry (VO) and visual simultaneous localization and mapping (vSLAM), the random sample consensus (RANSAC) \cite{fischlerRandomSampleConsensus1981} is the most commonly used algorithm.
RANSAC is an iterative algorithm to estimate a model from multiple observations which include outliers.
Thus, common vision-based methods often adopt this algorithm to enhance their performance in dynamic environments.
\add{
One of the major drawbacks is that it is more likely to fail when the observation contains outliers that follow the same hypothesis.
For example, RANSAC in visual localization is more vulnerable to one large rigid-body object than several small objects with different motions.
}

Some recent works report that the use of semantic segmentation or object detection can improve the robustness of algorithms in dynamic environments.
Mask-SLAM \cite{kanekoMaskSLAMRobustFeatureBased2018} uses the ORB-SLAM architecture while masking out dynamic objects using a mask generated from semantic segmentation.
DynaSLAM \cite{bescosDynaSLAMTrackingMapping2018} is built on ORB-SLAM2 and combines a geometric approach with semantic segmentation to remove dynamic objects.
The authors of Mask-SLAM and DynaSLAM evaluate proposal methods on their original dataset recorded in dynamic environments.
\add{
Empty Cities \cite{bescosEmptyCitiesDynamicObjectInvariant2020} integrates dynamic object detection with a generative adversarial model to inpaint the dynamic objects and generate static scenes from images in dynamic environments. 
}

\add{
There are several works that not only address robust localization in dynamic environments but also extend the capability to track surrounding objects \cite{bescosDynaSLAMIITightlyCoupled2020, qiuTracking3DMotion2019, sabzevariMultibodyMotionEstimation2016, juddMultimotionVisualOdometry2018}.
For example, the results of DynaSLAM II \cite{bescosDynaSLAMIITightlyCoupled2020} show that the objects motion estimation can be beneficial for ego-motion estimation in dynamic environments.
}


Compared to VO/vSLAM, the performance of VIO algorithms in dynamic environments has not been investigated in depth.
\unnecessary{
\cite{zhuChallengesDynamicEnvironment2018} reported a comprehensive analysis of the impact of noisy measurements on the accuracy and robustness of VIO. 
These noisy measurements include moving objects, but they did not provide how they evaluated the impact of these objects on VIO.
}
\add{Several works, however, report that the utilization of semantic/instance segmentation improves the performance of VIO in dynamic environments \cite{muVisualNavigationFeatures2020, baiPerceptionaidedVisualInertialIntegrated2020}.}
A limitation of these two studies is, however, that they reported performances only on limited data.
For example, \cite{muVisualNavigationFeatures2020} uses only a short subsequence of the KITTI dataset.
Our VIODE benchmark is well suited to systematically assess and compare these VIO algorithms.
\section{The VIODE Dataset}\label{section:dataset}

We created the VIODE dataset with AirSim~\cite{shahAirSimHighFidelityVisual2017}, a photorealistic simulator based on Unreal Engine 4, which features commonly used sensors such as RGB cameras, depth cameras, IMUs, barometers, and LiDARs.
To simulate motions that are typically encountered in VIO applications, we used a quadrotor as the carrier vehicle as it is a platform where VIO algorithms are commonly employed.

The VIODE dataset is designed for benchmarking VIO algorithms in dynamic environments. 
This can be accredited to three unique features.
Firstly, compared to other existing datasets, VIODE contains highly dynamic sequences.
Secondly, we provide quantitative measures of the dynamic level of the sequences based on clearly defined metrics.
Finally, and most importantly, VIODE allows us to isolate the effect of dynamic objects on the robustness of VIO.
In general, the performance of VIO is influenced not only by dynamic objects but also by a variety of parameters such as the accuracy of the IMU, the type of movements (e.g. rotation or translation), the lighting condition, and the texture of surrounding objects.
Therefore, a way to fairly assess algorithm robustness in dynamic scenarios is to compare multiple sequences where the only difference is the dynamic level.
To achieve this, we create four sequences of data on the same trajectory and strategically add dynamic objects.

\subsection{Dataset content}
\begin{figure}[tbp]
    \centering
    \vspace{0.3cm}
    \includegraphics[width=\columnwidth]{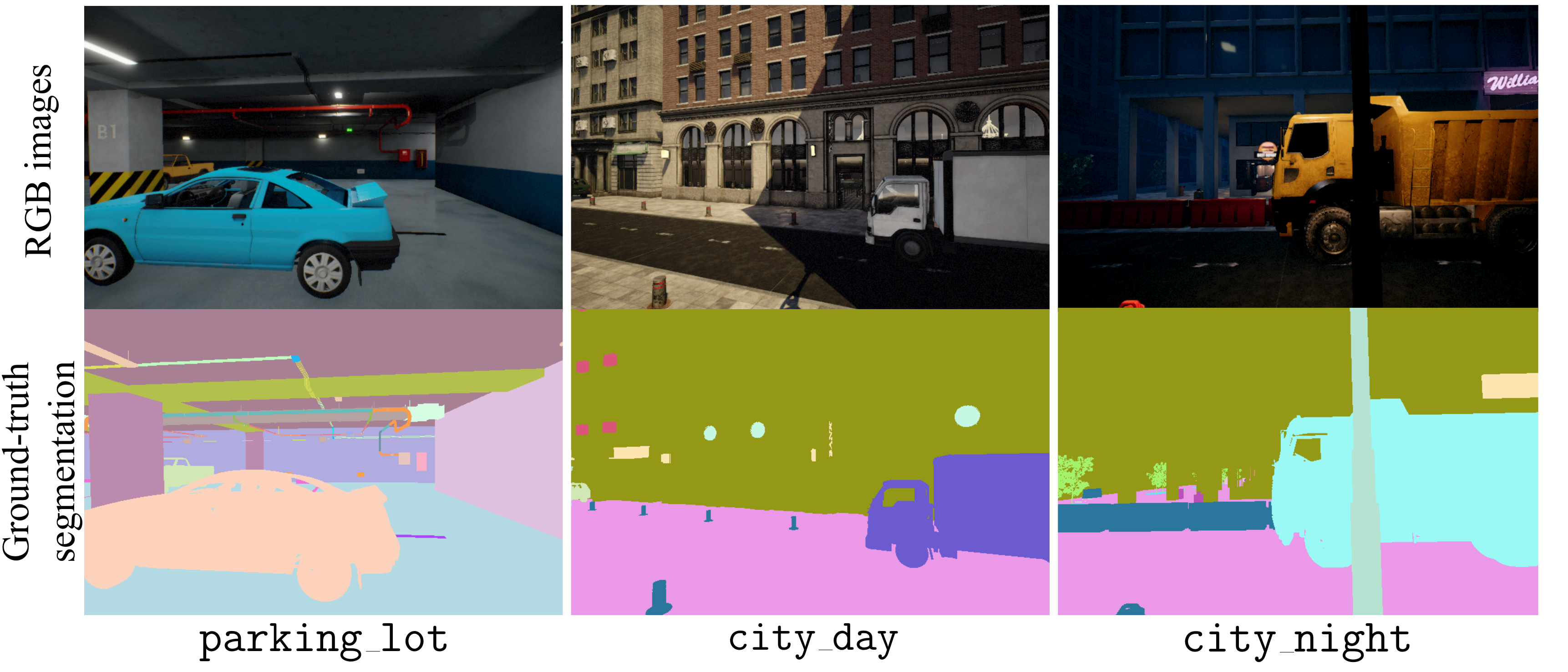}
    \caption{\footnotesize The images in the first row are RGB image examples from each of the environments and the images in the second row are the corresponding ground-truth segmentation maps provided by AirSim. In the VIODE dataset, we included these two images in a time-synchronization fashion.}
    \label{fig:proposed_dataset_overview}
\end{figure}

The dataset is recorded in three \textit{environments}, one indoor and two outdoor environments.
In each environment, we generate four \textit{sequences} which are recorded while executing the same trajectory. 
The only difference between these four recorded sequences is the number of dynamic vehicles.
To set up the situation as realistic as possible, we also placed static vehicles in all the sequences.
The three environments in our dataset are as follows (see \figref{fig:proposed_dataset_overview}):
\begin{itemize}
    \item \texttt{parking\_lot} contains sequences in parking lot indoor environment. 
    \item \texttt{city\_day} contains sequences during daytime in modern city environment. This is a commonly encountered outdoor environment which is surrounded by tall buildings and trees.
    \item \texttt{city\_night} contains sequences during night time in the same environment as \texttt{city\_day}. Trajectories of ego-vehicle and dynamic objects are different from those of the \texttt{city\_day} environment.
\end{itemize}

\subsection{Dataset generation}
The procedure for generating VIODE data involves two steps: 1. collect IMU and ground-truth 6-DoF poses, and 2. capture images and segmentation masks (see \figref{fig:catchy}). 
These steps were executed on a desktop computer running Windows 10 with an Intel(R) Core i7-9700 CPU and a GeForce RTX 2700 SUPER GPU. 

In the first step, we use the ROS wrapper provided by AirSim to generate synchronized IMU measurements and ground-truth odometry data at a rate of 200~Hz.
While recording this sensory data, we control the UAV with the PythonAPI in AirSim.
We generate one trajectory for each environment.
Each of the trajectories contains linear accelerations along the x/y/z axes and rotation movements around roll/pitch/yaw.
In order to obtain accurate timestamps, we slow down the simulation by a factor of 0.05 during this procedure.

In a second step, we add images to the above data.
To do so, we undersample ground-truth poses from the odometry sequence to a rate of 20~Hz. 
In each of the undersampled poses, we capture synchronized RGB images and segmentation maps with stereo cameras \unnecessary{with PythonAPI}. 
This procedure is done four times in each environment to generate four sequences with different dynamic levels: \texttt{none}, \texttt{low}, \texttt{mid}, and \texttt{high}.

\subsection{Sensor setup and calibration}
The UAV in AirSim is equipped with a camera and an IMU.
The camera captures RGB images and AirSim provides segmentation maps.

\begin{itemize}
    \item {\bf Stereo camera} uses two equal cameras and captures time-synchronized RGB images at a rate of 20~Hz with WVGA resolution ($752\times480$ px). We set up the stereo camera with a baseline of 5~cm. Both of these cameras are global shutter cameras and have a $90^\circ$ field of view (FOV). 
    \item {\bf Ground-truth segmentation} images are computed for both cameras at a rate of 20~Hz, synchronized with the corresponding RGB images (see \figref{fig:proposed_dataset_overview}). The provided segmentation is an instance segmentation. As such, the vehicles and all the other objects in the scene are labeled as different objects.
    \item {\bf IMU} data is also acquired from AirSim at a rate of 200~Hz. This uses the same parameters as MPU-6000 from TDK InvenSense. The noise follows the model defined in \cite{woodmanIntroductionInertialNavigation2007}. Since our IMU data does not take into account the vibration of rotors, the noise magnitude is lower than that of IMU data recorded on real-world UAV flights and similar to the one recorded with handheld carriers.
    \item {\bf Ground-truth trajectory} is also included for all the sequences in our dataset. This consists of ground-truth 6-DoF poses provided by the simulator. This is recorded at a rate of 200~Hz.
    \item We also provide ground-truth {\bf intrinsic and extrinsic parameters} for the stereo camera and camera-IMU extrinsics.
\end{itemize}

\begin{table*}
\vspace{0.5cm}
\centering
\caption{\textsc{basic parameters of viode dataset}}
\label{tab:dataset}
\begin{tabular}{l>{\raggedright}p{0.06\linewidth}>{\raggedright\arraybackslash}p{0.05\linewidth}>{\raggedright\arraybackslash}p{0.05\linewidth}>{\raggedright\arraybackslash}p{0.05\linewidth}>{\raggedright\arraybackslash}p{0.05\linewidth}>{\raggedright\arraybackslash}p{0.06\linewidth}>{\raggedright\arraybackslash}p{0.08\linewidth}>{\raggedright\arraybackslash}p{0.08\linewidth}>{\raggedright\arraybackslash}p{0.07\linewidth}>{\raggedright\arraybackslash}p{0.07\linewidth}}
\bhline{0.8pt}
& & Indoor/ outdoor &  Distance {[}m{]} & Duration {[}s{]} & \# of static objects & \# of dynamic objects & Average pixel-based dynamic rate~{[}\%{]} & Maximum pixel-based dynamic rate~{[}\%{]} & Average OF-based dynamic rate~{[}px{]} & Maximum OF-based dynamic rate~{[}px{]}\\ \hline
\multicolumn{1}{l}{\multirow{4}{*}{\texttt{parking\_lot}}} & {\texttt{none}}~~~~~~~ & \multirow{4}{*}{Indoor} &\multirow{4}{*}{75.8} & \multirow{4}{*}{59.6} & \multirow{4}{*}{2} & 0 & 0.0 & 0.0 & 0.0 & 0.0\\
\multicolumn{1}{l}{} & {\texttt{low}} & & & & & 1 & 0.5 & 13.1 & 0.03 & 0.83\\
\multicolumn{1}{l}{} & {\texttt{mid}} & & & & & 2 & 8.9 & 68.6 & 0.22 & 3.92\\
\multicolumn{1}{l}{} & {\texttt{high}} & & & & & 6 & 10.5 & 66.9 & 0.31 & 3.75\\ \hline
\multicolumn{1}{l}{\multirow{4}{*}{\texttt{city\_day}}} & {\texttt{none}} & \multirow{4}{*}{Outdoor} & \multirow{4}{*}{157.7} & \multirow{4}{*}{66.4} & \multirow{4}{*}{2} & 0 & 0.0 & 0.0 & 0.0 & 0.0\\
\multicolumn{1}{l}{} & {\texttt{low}} & & & & & 1 & 1.1 & 50.3 & 0.03 & 4.66 \\
\multicolumn{1}{l}{} & {\texttt{mid}} & & & & & 3 & 2.0 & 52.1 & 0.14 & 5.51 \\
\multicolumn{1}{l}{} & {\texttt{high}} & & & & & 11 & 8.4 & 98.8 & 0.45 & 8.72\\ \hline
\multicolumn{1}{l}{\multirow{4}{*}{\texttt{city\_night}}} & {\texttt{none}} & \multirow{4}{*}{Outdoor} & \multirow{4}{*}{165.7} & \multirow{4}{*}{61.6} & \multirow{4}{*}{1} & 0 & 0.0 & 0.0 & 0.0 & 0.0\\
\multicolumn{1}{l}{} & {\texttt{low}} & & & & & 1 & 0.9 & 33.4 & 0.08 & 3.10\\
\multicolumn{1}{l}{} & {\texttt{mid}} & & & & & 3 & 1.6 & 32.9 & 0.11 & 3.59\\
\multicolumn{1}{l}{} & {\texttt{high}} & & & & & 11 & 3.5 & 37.6 & 0.21 & 3.31\\ \bhline{0.8pt}
\end{tabular}
\end{table*}

\subsection{Dynamic objects \& dynamic level evaluation}
We use cars provided in Unreal Engine 4 as dynamic objects since cars are commonly encountered dynamic objects in real-world applications.
We use six types of vehicles with varying textures and sizes.
These dynamic objects are controlled by Unreal Engine with constant speeds along the designated trajectories.
As mentioned previously, we generated four sequences in each environment: \texttt{none}, \texttt{low}, \texttt{mid}, and \texttt{high}.
The \texttt{none} sequences do not contain any dynamic objects.
Starting from \texttt{none}, we progressively add the vehicles to generate three different dynamic sequences: \texttt{low}, \texttt{mid}, and \texttt{high}.
Furthermore, we allocate not only dynamic vehicles but also some static ones to make the scenes more realistic.

We also introduce metrics to evaluate how dynamic the sequences are.
Simply counting the number of dynamic vehicles in the scene is not a suitable metric of the dynamic level.
For example, even if the number of dynamic objects is the same, the scene is more dynamic when the objects are close to the camera.
Thus, a better metric is necessary to evaluate the dynamic level.
Possible information sources for defining a better metric are semantic information, optical flow, ground-truth ego-motion, or \add{rigid-body-motion of vehicles}. 
Here, we assess the dynamic level by considering how much of the FOV is occupied by dynamic objects based on ground-truth instance segmentation.
Pixel-based dynamic rate $r_{\rm{pix}}$ is defined as
\begin{align}\label{eq:dynamic_rate}
    {r_{\rm{pix}} = N^{\rm dyn}_{\rm pix} / N^{\rm all}_{\rm pix}}
\end{align}
where $N^{\rm dyn}_{\rm pix}$ is a number of pixels occupied by dynamic vehicles and $N^{\rm all}_{\rm pix}$ is the total number of pixels.

However, pixel-based dynamic rate does not contain speed information of the surrounding objects.
In order to obtain a metric for the magnitude of the motion of objects, we also define the optic-flow-based (OF-based) dynamic rate $r_{\rm{of}}$ as
\begin{align}\label{eq:of-based_dynamic_rate}
    r_{\rm{of}} = \frac{1}{N}\sum_{(x, y)\in D} {\sqrt{\| \nabla I(x, y) - \nabla I_{\rm{none}}(x, y) \|^2}}
\end{align}
where $D$ is a set of pixel coordinates in the frame which belong to dynamic objects.
We determine $D$ from ground-truth instance segmentation.
$N$ is the total number of pixels, $\nabla I(x, y)$ is the optic flow at $(x, y)$ in a frame, and $\nabla I_{\rm{none}}(x, y)$ is the optic flow at $(x, y)$ in a corresponding frame in {\texttt{none}} sequence from the same environment.


We summarize these parameters as well as the basic parameters such as the total distance and duration of each data sequence (see \tabref{tab:dataset}).
We further provide the two types of dynamic rate along the time axis (see \figref{fig:eval_errorVSdynrate}).
\section{Evaluation of VIO Methods On The VIODE Dataset}
\subsection{Evaluation metrics}
To analyze the performance of VIO algorithms, we use the absolute trajectory error (ATE)~\cite{sturmBenchmarkEvaluationRGBD2012a} which is defined as
\add{
\begin{align}
    {\rm{ATE}}(\mathbf{P}_{1:n}, \mathbf{Q}_{1:n}) = \sqrt{\frac{1}{n} \sum_{i=1}^{n} \|\text{trans}(\mathbf{F}_i)\|^2}
\end{align}
where $\mathbf{P}_1, ..., \mathbf{P}_n \in \rm{SE}(3)$ is the estimated trajectory, $\mathbf{Q}_1, ..., \mathbf{Q}_n \in \rm{SE}(3)$ the ground-truth trajectory, $\mathbf{F}_i = \mathbf{Q}_i^{-1}\mathbf{S}\mathbf{P}_i$ is the absolute trajectory error at time step $i$, $\text{trans}(\mathbf{X})$ refers to the translational components of $\mathbf{X} \in \rm{SE}(3)$, and $\mathbf{S}$ the rigid-body transformation between $\mathbf{P}_{1:n}$ and $\mathbf{Q}_{1:n}$ calculated by optimizing a least-square problem.
Moreover, we use relative pose error (RPE)~\cite{sturmBenchmarkEvaluationRGBD2012a} for local accuracy of the sub-trajectory.
RPE at $i$-th frame is defined as
\begin{align} \label{eq:subtraj_ate}
    {\rm{RPE}}(\mathbf{P}_{1:n}, \mathbf{Q}_{1:n}, i) = \|\text{trans}(\mathbf{E}_i)\|
\end{align}
where $\mathbf{E}_i = (\mathbf{Q}_i^{-1}\mathbf{Q}_{i + \Delta})^{-1}(\mathbf{P}_i^{-1}\mathbf{P}_{i + \Delta})$ is relative pose error at time step $i$.
Here, $\Delta$ is a fixed time interval for which we calculate the local accuracy.
}

We additionally introduce degradation rate $r_d$ for the VIODE dataset.
This is defined as \inlineequation[eq:degradation_rate]{r_d={\rm ATE}_{\rm high}/{\rm ATE}_{\rm none}}, where ${\rm ATE}_{\rm high}$ and ${\rm ATE}_{\rm none}$ are the ATE of the algorithm in the \texttt{high} and \texttt{none} sequences respectively.
This metric illustrates the robustness of the VIO algorithm in dynamic sequences compared to that in static sequences.
We calculate this value for all evaluated algorithm in each environment.

\subsection{Existing VIO algorithms}
\begin{table}[tb]
    \renewcommand\arraystretch{1.3}
    \centering
    \caption{\textsc{degradation rate $r_d$ for each algorithm/environment}}
    \begin{tabular}{llll}
    \bhline{0.8pt}
    Environment\textbackslash Algorithm & ROVIO & VINS-Mono & VINS-Mask (Ours) \\ \hline
    {\texttt{parking\_lot}} & 5.27 & 14.7 & \bf{1.12}\\
    {\texttt{city\_day}} & 17.1 & 16.2 & \bf{1.26} \\
    {\texttt{city\_night}} & 5.19 & 1.91 & \bf{1.26}\\\bhline{0.8pt}
    \end{tabular}
    \label{tab:relative_error}
\end{table}

We apply ROVIO and VINS-Mono, two state-of-the-art VIO algorithms, on the VIODE dataset.
Since the performance of these algorithms is not deterministic, each is evaluated ten times.

\begin{figure*}[t]
    \centering
    \includegraphics[width=2\columnwidth]{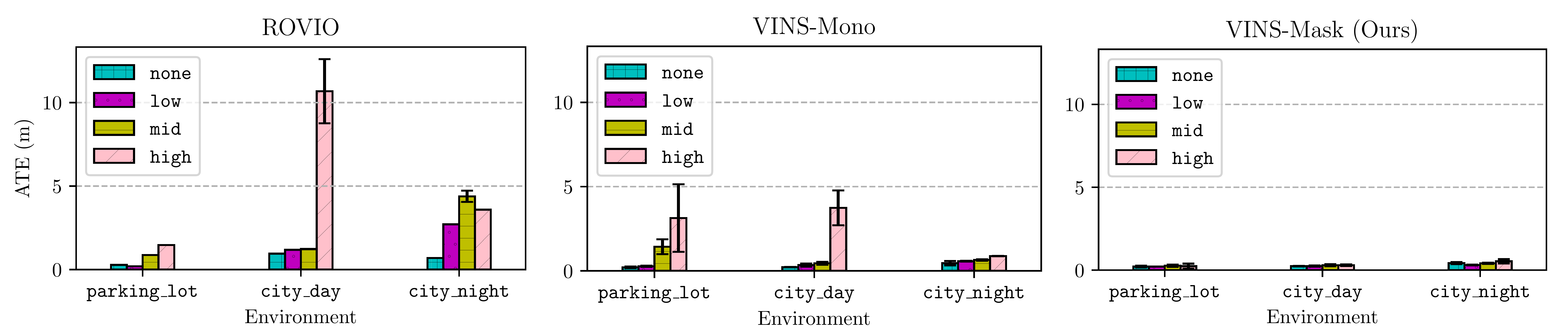}
\caption{\footnotesize Absolute trajectory error (ATE) [m] of ROVIO, VINS-Mono, and VINS-Mask. The performance of these methods is not deterministic. Thus, we run the simulation ten times for every sequence. We observe an increase in errors as the dynamic level increases for ROVIO and VINS-Mono. On top of that, the ATE of VINS-Mask remains considerably lower than the other two algorithms.}
    \label{fig:eval_error}
\end{figure*}

\begin{figure*}[htb]
\vspace{0.15in}
    \centering
    \includegraphics[width=2.0\columnwidth]{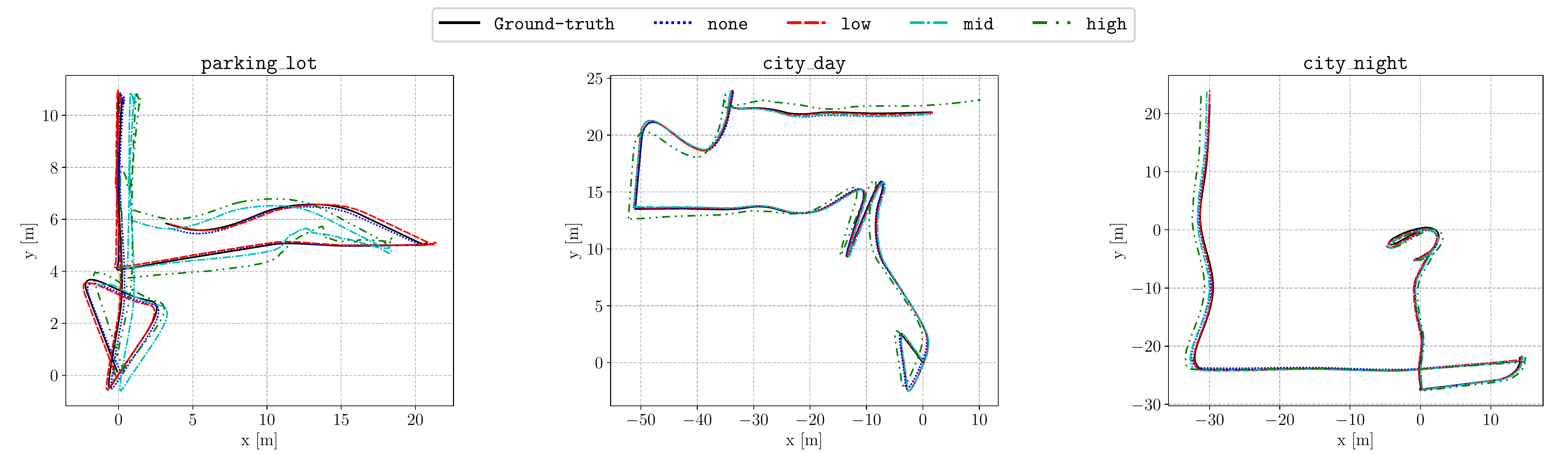}
\caption{\footnotesize Estimated trajectory of VINS-Mono in \texttt{none}, \texttt{low}, \texttt{mid}, and \texttt{high} sequence for each environment.}
    \label{fig:eval_traj_dynamic}
\end{figure*}

Our findings show that both ROVIO and VINS-Mono perform worse in the presence of dynamic objects.
We observe that the ATE of both algorithms increases as the scene gets more dynamic (see \figref{fig:eval_error}).
Furthermore, estimated trajectories illustrate the degradation of VINS-Mono in highly-dynamic sequences.
For example, in \texttt{parking\_lot/high}, there is a drift around $(x, y)=(0, 8)$ while is not present in \texttt{parking\_lot/none} (see \figref{fig:eval_traj_dynamic}).
The degradation rate is mostly higher than 5.0 for ROVIO and VINS-Mono (see \tabref{tab:relative_error}).
Comparing the degradation rate of ROVIO and VINS-Mono among the three environments, {\texttt{city\_day}} environment is the most challenging one among the three.
The degradation rate of both ROVIO and VINS-Mono was the lowest in {\texttt{city\_night}}, in which the average and maximum pixel-based dynamic rate were also the lowest (see \figref{fig:proposed_dataset_overview}).

\begin{figure*}[t]
    \centering
    \includegraphics[width=1.96\columnwidth]{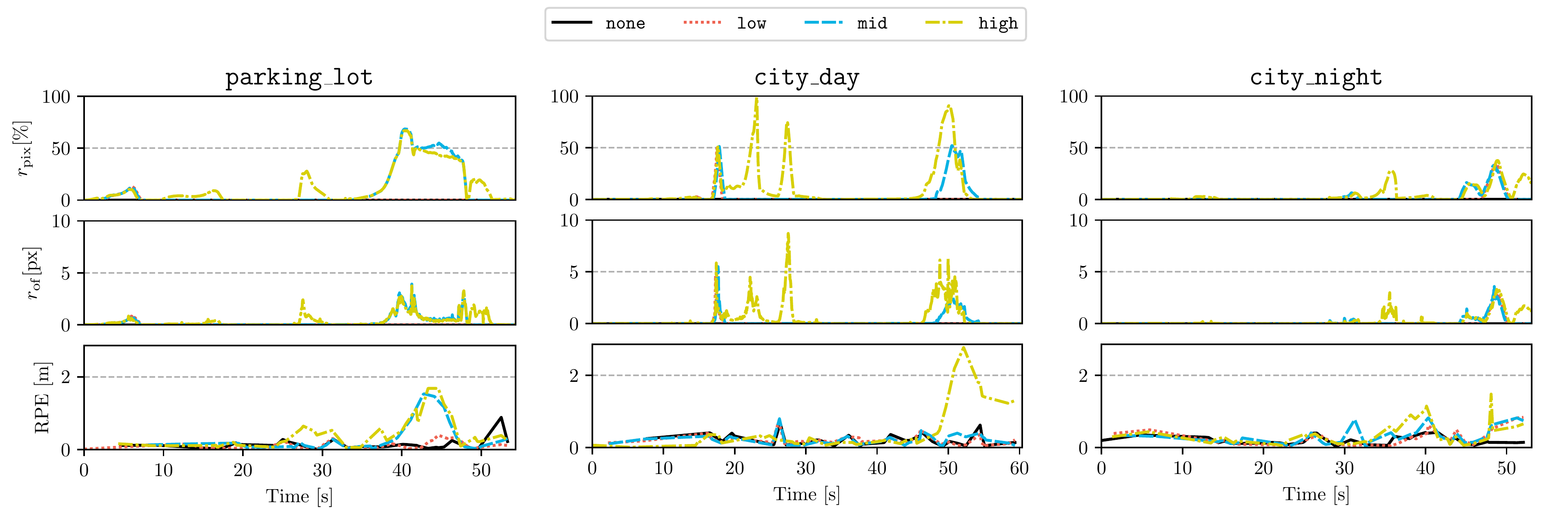}
    \caption{\footnotesize Comparison of two types of dynamic rate and estimation performance by VINS-Mono along time axis for \texttt{none/low/mid/high} sequences in each environment. First row is the pixel-based dynamic rate $r_{\rm{pix}}$ (see Eq.~\ref{eq:dynamic_rate}), second row is the OF-based dynamic rate $r_{\rm{of}}$ (see Eq.~\ref{eq:of-based_dynamic_rate}), and the last row is the \add{RPE} (see Eq.~\ref{eq:subtraj_ate}).}
    \label{fig:eval_errorVSdynrate}
\end{figure*}

We show that the wrong estimation often occurs when the dynamic objects are in FOV.
One evidence is a correspondence between two types of dynamic rate ($r_{\rm{pix}}$ and $r_{\rm{of}}$) and \add{RPE} estimated by VINS-Mono (see \figref{fig:eval_errorVSdynrate}).
For example, $r_{\rm{pix}}$, $r_{\rm{of}}$, and \add{RPE} mark the highest value during $40-50$~s in {\texttt{parking\_lot/high}}.
These correspondences between \add{RPE} and two types of dynamic rate support our hypothesis that the performance degradation of current VIO algorithms is caused by dynamic objects.

\subsection{VINS-Mask}
\begin{figure}[tb]
    \centering
    \includegraphics[width=\columnwidth]{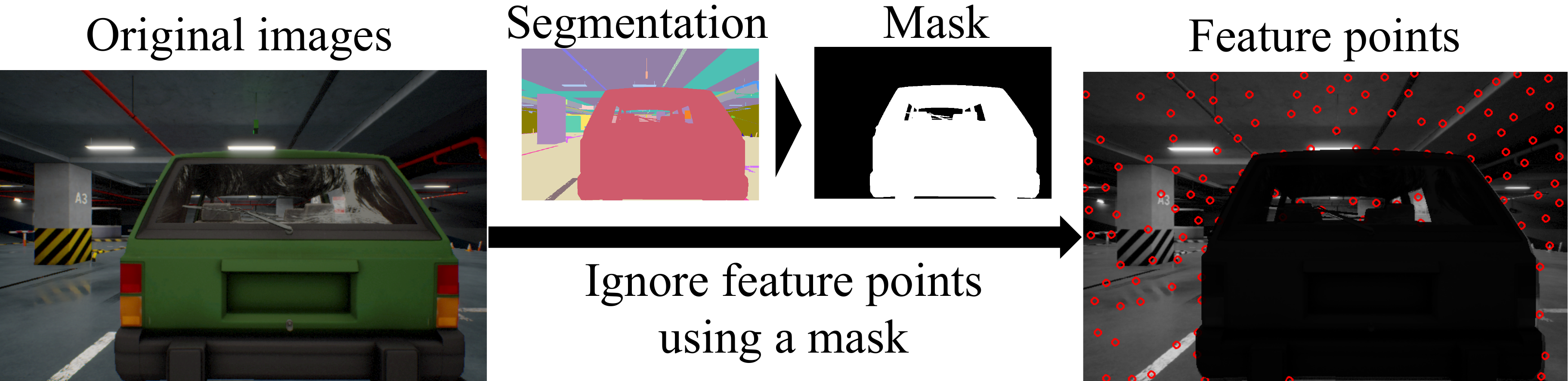}
    \caption{\footnotesize In VINS-Mask, we create a binary mask from segmentation maps to ignore dynamic regions. Since this is based on VINS-Mono, these feature points are coupled with IMU information to estimate the 6-DoF poses.}
    \label{fig:overview_vins_mask}
\end{figure}
\begin{figure}[tb]
    \centering
    \includegraphics[width=\columnwidth]{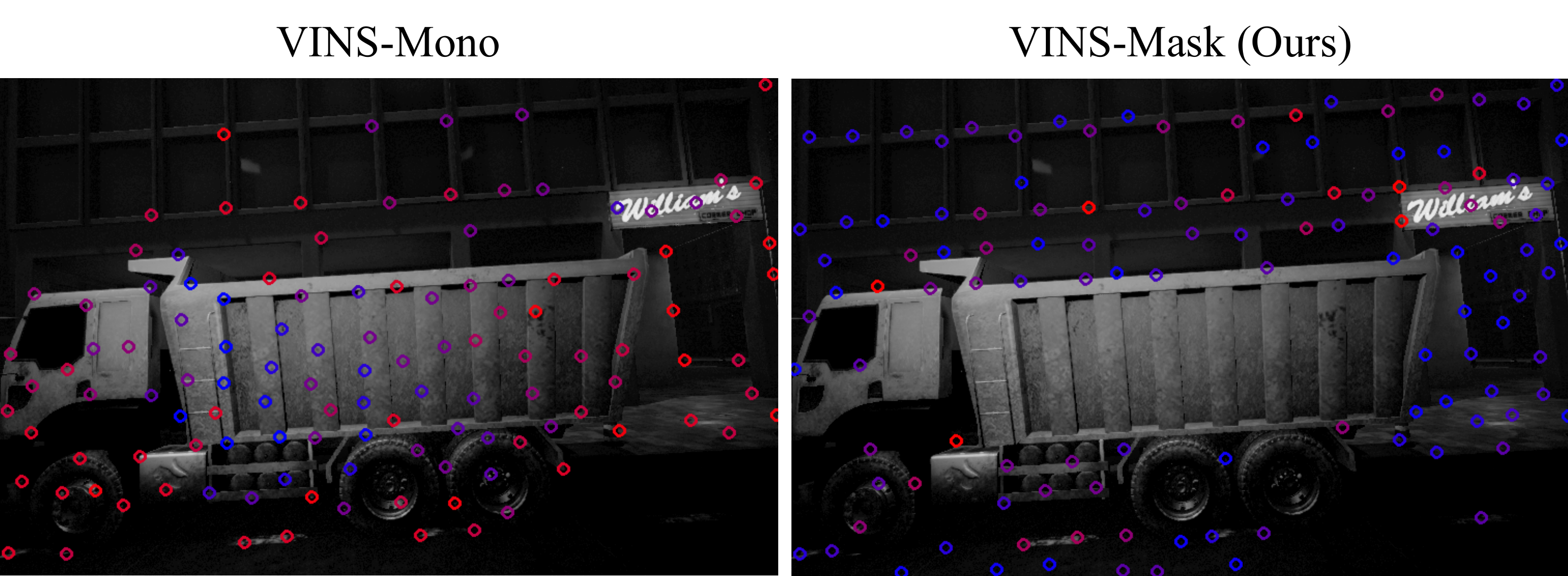}
    \caption{\footnotesize We illustrate the extracted and tracked feature points in VINS-Mono and VINS-Mask as red and blue circles. The more red the feature point is colored, the longer the feature has been tracked and more likely to be used in VINS-Mono \& VINS-Mask to estimate 6-DoF pose. VINS-Mono relies a lot on feature points on dynamic objects while VINS-Mask excludes those regions.}
    \label{fig:eval_feature_points}
\end{figure}
One of our goals is to show that scene semantics have the potential to improve the performance of VIO in challenging dynamic scenes.
To this end, we develop the method VINS-Mask, which is based on VINS-Mono \cite{qinVINSMonoRobustVersatile2018}.
VINS-Mask uses the same algorithm as VINS-Mono, except that it avoids using feature points on dynamic objects by leveraging semantic information.
In the feature points extraction phase of VINS-Mask, we perform semantic/instance segmentation for each image. 
By exploiting the segmentation and preliminary knowledge, we mask out the region of the dynamic objects right before the feature extraction phase of VINS-Mono (see \figref{fig:overview_vins_mask}).
By applying the mask, VINS-Mask can estimate the robots' pose based on reliable feature points (see \figref{fig:eval_feature_points}).
Although this technique is not a novel idea as stated in Sec.~\ref{section:dataset}, we use this method to assess the impact of semantic information on the robustness of VIO in dynamic scenarios.

In our work, we use ground-truth segmentation labels provided in the VIODE dataset, instead of applying a semantic segmentation algorithm on camera images.
The mask consists of the region which belongs to vehicles.
We include both dynamic and static vehicles in the mask to make the setup more realistic, although it is possible to distinguish moving objects from static objects with ground-truth instance segmentation.
This is on account of the fact that it is another challenging task to distinguish moving objects from static objects in real-world applications.

\begin{figure*}[htb]
    \centering
    \includegraphics[width=2.0\columnwidth]{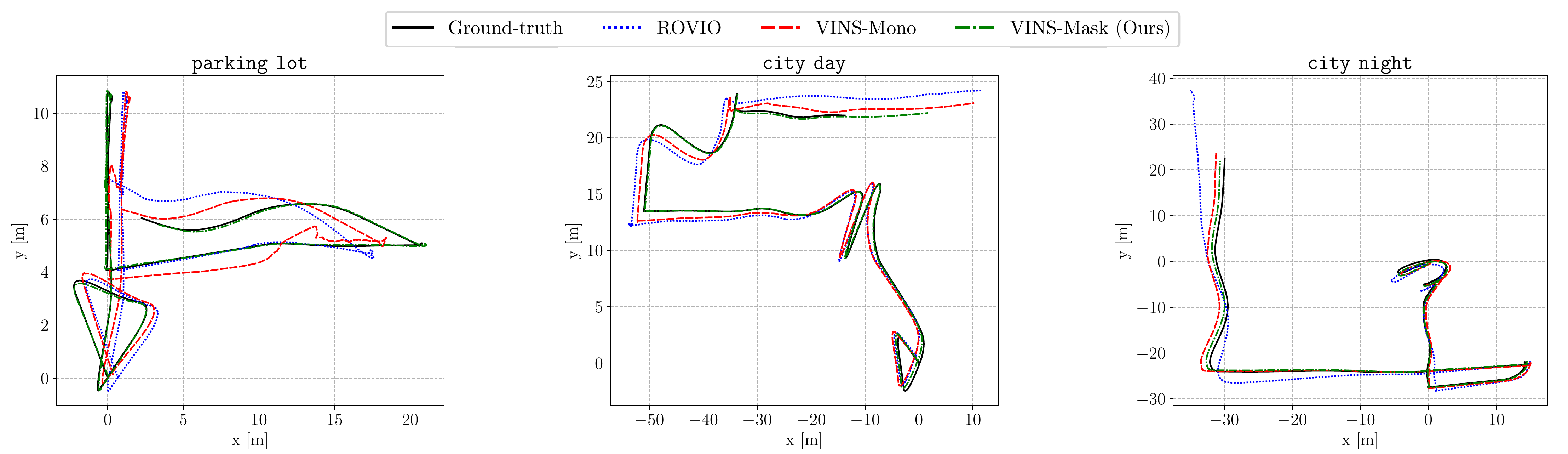}
    \caption{\footnotesize Estimated trajectory of ROVIO, VINS-Mono, and VINS-Mask with ground-truth for \texttt{high} sequences.}
    \label{fig:eval_traj_methods}
\end{figure*}

In addition to ROVIO and VINS-Mono, here we evaluate VINS-Mask on the VIODE dataset.
VINS-Mask is also applied ten times for each sequence.
We found that the performance of the VINS-Mask is almost independent of the dynamic level.
The ATE of VINS-Mask is mostly lower than the results of ROVIO and VINS-Mono (see \figref{fig:eval_error}).
The degradation rate of VINS-Mask is consistently lower than the other two algorithms (see \tabref{tab:relative_error}).
The estimated trajectories in \texttt{high} sequences for each algorithm also \unnecessary{clearly} illustrates the improvement by VINS-Mask (see \figref{fig:eval_traj_methods}).
While the existing algorithms exhibit some drifts in highly dynamic sequences, VINS-Mask successfully suppresses these drifts.
It is also worth noting that VINS-Mask performs similarly with VINS-Mono in static sequences, in which VINS-Mask masks out static objects while VINS-Mono does not. 
The improvement by VINS-Mask suggests the usefulness of this type of approach as a technique to run VIO robustly in dynamic environments.
Besides that, the results further strengthen our confidence that the performance degradation of the existing algorithms is due to the dynamic objects in the scene.

\section{Conclusion}
In this paper, we proposed the VIODE dataset, a novel visual-inertial benchmark that contains variable and measurable dynamic events.
Through the systematic introduction of dynamic objects, the users can isolate the effect of dynamic objects on the robustness of VIO.
Using the VIODE dataset, we have shown that both VINS-Mono and ROVIO, the two state-of-the-art open-source VIO algorithms, perform worse as the scenes get more dynamic.
We also demonstrated that the utilization of semantic information has the potential to overcome this degradation.
By masking out the region of the objects, we could mitigate the impact of dynamic objects on the VIO algorithm.

As a future research direction, it is necessary to evaluate the VINS-Mask by using semantic segmentation instead of ground-truth segmentation.
One of the challenges would be a real-time onboard deployment, as latency can be critical for the performance of VIO.
It is also important to investigate the better utilization of semantic segmentation for VIO.
One of the directions would be to distinguish static objects from dynamic ones.
Since the current VINS-Mask can mask outs static objects such as parked vehicles, further research is still necessary for robust VIO in challenging dynamic scenes.
\add{
As a future direction of the VIODE dataset, we are also interested in introducing other types of sensors in the dataset to enable the evaluation of localization algorithms with various sensor configurations.
Another possible future work is, similarly to \cite{antoniniBlackbirdUAVDataset2020a}, to use real IMU and trajectory data recorded on a real UAV platform instead of simulated ones.
}

\section*{Acknowledgments}

This research was partially supported by the Swiss National Science Foundation (SNF) with grant number 200021-155907 and the Swiss National Center of Competence in Research (NCCR).



\bibliographystyle{IEEEtran}
\bibliography{000_root}

\end{document}